\documentclass[11pt,a4paper]{article}
\usepackage[hyperref]{acl2020}
\usepackage{times}
\usepackage{latexsym}

\usepackage{graphicx}
\usepackage{tabularx}
\usepackage{xcolor, soul}
\colorlet{ct1}{green!66.55}
\colorlet{ct2}{green!53.25}
\colorlet{ct3}{green!52.29}

\colorlet{at1}{green!71.06}
\colorlet{at2}{green!26.82}
\colorlet{at3}{green!12.09}
\colorlet{at4}{green!55.66}
\colorlet{at5}{green!13.47}
\colorlet{at6}{green!22.14}
\colorlet{at7}{green!17.73}

\colorlet{co1}{green!19.39}
\colorlet{co2}{green!32.98}
\colorlet{co3}{green!20.94}
\colorlet{co4}{green!72.14}
\colorlet{co5}{green!23.51}
\colorlet{co6}{green!42.08}
\colorlet{co7}{green!23.86}

\colorlet{ao1}{green!54.96}
\colorlet{ao2}{green!21.50}
\colorlet{ao3}{green!48.38}
\colorlet{ao4}{green!11.98}
\colorlet{ao5}{green!18.67}
\colorlet{ao6}{green!23.28}
\colorlet{ao7}{green!24.58}
\colorlet{ao8}{green!21.57}
\colorlet{ao9}{green!22.85}
\colorlet{ao10}{green!26.41}
\colorlet{ao11}{green!28.22}
\usepackage{microtype}

\aclfinalcopy 
\def\aclpaperid{15} 

\title{Exploring Weaknesses of {VQA} Models through Attribution Driven Insights
}

\author{Shaunak Halbe \\
  College of Engineering Pune \\
  \texttt{shaunak9@ieee.org} \\
 }

\date{}

\begin{document}
\maketitle
\begin{abstract}
Deep Neural Networks have been successfully used for
the task of Visual Question Answering for the past few years
owing to the availability of relevant large scale datasets.
However these datasets are created in artificial settings and
rarely reflect the real world scenario. Recent research effectively applies these {VQA} models for answering visual questions for the blind. Despite achieving high accuracy these
models appear to be susceptible to variation in input questions.We analyze popular {VQA} models through the lens of
attribution (input's influence on predictions) to gain valuable insights. Further, We use these insights to craft adversarial attacks which inflict significant damage to these systems with negligible change in meaning of the input questions. We believe this will enhance development of systems
more robust to the possible variations in inputs when deployed to assist the visually impaired.

\end{abstract}

\section{Introduction}
Visual Question Answering (VQA) is a semantic task,
where a model attempts to answer a natural language question based on the visual context. With the emergence of
large scale datasets \cite{VQA,balanced_vqa_v2,krishnavisualgenome,malinowski2014nips,zhu2016visual7w}, There has been outstanding
progress in VQA systems in terms of accuracy obtained on
the associated test sets. However these systems are seen to
somewhat fail when applied in real-world situations \cite{gurari2018vizwiz,agrawal-etal-2016-analyzing}
majorly due to a significant domain shift and an inherent
language/image bias. A direct application of VQA is to
answer the questions for images captured by blind people.
The VizWiz \cite{gurari2018vizwiz} is a first of its kind goal oriented dataset
which reflects the challenges conventional VQA models
might face when applied to assist the blind. The questions
in this dataset are not straightforward and are often conversational which is natural knowing that they have been asked by visually impaired people for assistance. Due to unsuitable images or irrelevant questions most of these questions are unanswerable. These questions differ from those in other datasets mainly in the type of answer they are expecting. The questions are often subjective and require the algorithm to actually read (OCR)/ detect/ count, moreover understand the image before answering. We believe models trained on such a challenging dataset must be interpretable and should be analyzed for robustness to ensure they are accurate for the right reasons.
\section{Model Interpretability}
Deep Neural Networks often lack interpretability but are
widely used owing to their high accuracy on the representative test sets. In most applications a high test-set accuracy is sufficient, but in certain sensitive areas, understanding causality is crucial. When deploying such VQA models to aid the blind, utmost care needs to be taken to prevent the model from answering wrongly to avoid possible accidents. In the past, various saliency methods have been used to interpret models which have textual inputs. Vanilla Gradient Method\cite{simonyan2013deep} visualizes the gradients of the loss with respect to each input token(word in this case). SmoothGrad \cite{smilkov2017smoothgrad} averages the gradient by adding Gaussian noise to the input. Layerwise Relevance Propagation (LRP) \cite{binder2016layer}, DeepLift \cite{shrikumar2017learning} are similar methods used for this purpose.
\section{Integrated Gradients (IG) }
Vanilla, LRP and DeepLift violate the axioms of Sensitivity and Implementational Invariance as discussed by \citealt{10.5555/3305890.3306024}. As Integrated Gradients (IG)\cite{10.5555/3305890.3306024} satisfies the necessary axioms, we use it for the purpose of interpretability. IG computes attributions for the input features based on the network's predictions. These attributions assign credit/blame to the input features (pixels in case of an image and words in case of a question) which are responsible for the output of the model. These attributions can help identify when a model is accurate for the wrong reasons like over-reliance on images or possible language priors. These attributions are computed with respect to a baseline input. In this paper, we use an empty question as the baseline. We use these attributions which specify word importance in the input question to design adversarial questions, which the model fails to answer correctly. While doing so, we try to preserve the original meaning of the question and ensure the simplicity of the same. We design these questions manually by incorporating highly attributed content-free words in the original question,taking into consideration the free-formed conversational nature of the questions that any user of such a system might ask. By content-free, we refer to words that are context independent like prepositions (e.g., "on", "in"), determiners (e.g., "this", "that") and certain qualifiers (e.g., "much", "many") among others.

\section{Related Work}
The main idea of adversarial attacks is to carefully perturb the input without making perceivable changes, in order to affect the prediction of the model. There has been significant research on adversarial attacks concerning images\cite{goodfellow2014explaining,madry2017towards}. These attacks exploit the oversensitivity of  models towards changes in the input image. \citealt{Sharma2018AttendAA} study attention guided implementations of popular image-based attacks on VQA models. \citealt{xu2018fooling} discuss methods to generate targeted attacks to perturb input images in a multimodal  setting. \citealt{ramakrishnan2018overcoming} observe that VQA models heavily rely on certain language priors to directly arrive at the answer irrespective of the image. They further develop a bias-reducing approach to improve performance. \citealt{kafle2017analysis} study the response of VQA models towards various question categories to indicate the deficiencies in the datasets. \citealt{huang2019novel} analyze the robustness of VQA models on basic questions ranked on the basis of similarity by LASSO based optimization method. Finally, \citealt{mudrakarta-etal-2018-model} use attributions to determine word importance and leverage them to craft adversarial questions. We adapt their ideas to the conversational aspect of questions in VizWiz to better suit our task. In this paper we restrict ourselves to attacks in the language domain, i.e. we only perturb the input questions and analyze the network's response.

\section{Robustness Analysis}

\subsection{Model and Data Specifications}
The VizWiz dataset \cite{gurari2018vizwiz} consists of 20,523 training
set image-question pairs and 4,319 validation pairs \cite{vizwiz_browser}.
Whereas the VQA v2 dataset \cite{balanced_vqa_v2} consists of 443,757
training questions and 214,354 validation questions. The
VizWiz dataset is significantly smaller than other VQA
datasets and hence is not ideal to determine word importance for the content free words. In order to do justice to
these words and to keep the analysis generalizable we use
the VQA v2 dataset for computing text attributions.
We use the Counter model \cite{zhang2018learning} for the purpose of computing attributions. This model is structurally similar to the
Q+I+A \cite{45997} (which was used to benchmark on VizWiz). We
select this model for ease in reproducibility and for consistency with the original paper \cite{gurari2018vizwiz}. We compute attributions
over the validation set, of which the highly attributed words
are selected to design prefix and suffix phrases which can
be incorporated in original questions for adversarial effect.Further we verify and test these attacks on the following
models : (1) Pythia \cite{singh2019pythia} (the VizWiz 2018 challenge winner) pretrained on VQA v2  and transferred to VizWiz
(train split) and (2) Q+I+A  model (which was used
to benchmark on VizWiz) trained from scratch on VizWiz
(train split).

\subsection{Observations}
We compute the total attribution that every word receives as well as average attribution for every word based on it's frequency of occurrence. We only take into account content free words, with the intention of preserving the meaning of the original question when these words are added to it. We observe that among the content-free words, "what",
"many", "is", "this" and "how" consistently receive high attribution in a question. We use these words along with some other context independent words to design the attacks.
We use these words to create seemingly natural phrases to be prepended or appended to the question. We observe that the model alters it's prediction under the influence of these added words.

\begin{figure}[t]
\begin{center}
\includegraphics[width=0.8\linewidth ,height = 5cm]{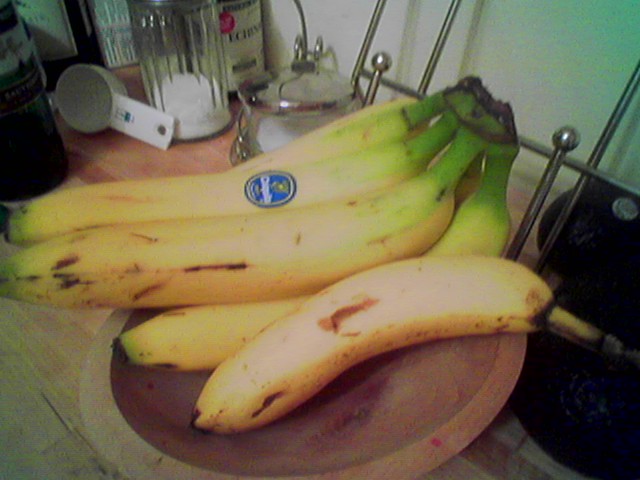}
\caption{Attributions overlaid on the corresponding input words. The output of the model changes from "yellow" to 1 which is driven by the word "many". }
\end{center}

\begin{center}
   \includegraphics[width=0.8\linewidth, height = 5cm]{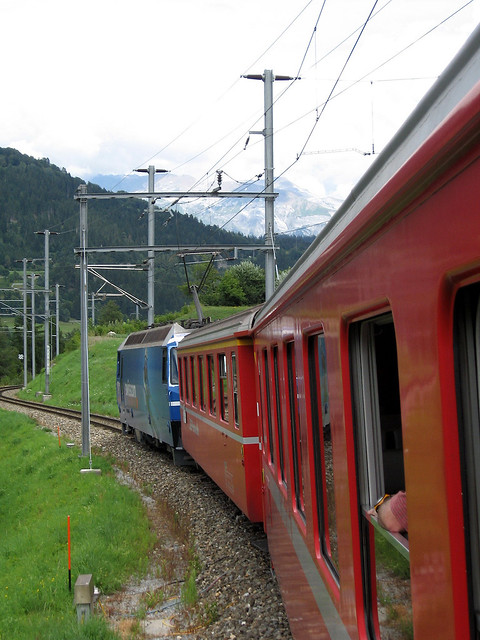}
   \caption{The output of the model is driven by the word "answer" acting as an adversary.}
   \end{center}
\label{fig:one}
\end{figure}

\subsection{Suffix Attacks}
We present Suffix Attacks, wherein we append content free phrases to the end of each question and evaluate the strength of these attacks through the accuracy obtained by the model on validation set and the percentage of answers it predicts as unanswerable/unsuitable (U). 

\subsection{Prefix Attacks}
We expand the Prefix attacks of  \citealt{mudrakarta-etal-2018-model} in a conversational vein to suit our task. These are seen to be more effective as prefix allows us to add important words like "What" and "How" to the start of a question which confuses the model to a greater extent than suffix attacks.  

\begin{table}[t]
\begin{center}
\begin{tabular}{c}
 Question : \\ \sethlcolor{co1} \hl{what}\sethlcolor{co2} \hl{is} \sethlcolor{co3}\hl{the} \sethlcolor{co4}\hl{color}
  \sethlcolor{co5}\hl{of} \sethlcolor{co6}\hl{this} \sethlcolor{co7}\hl{fruit} ? \\ \\ Predicted Label:\\ Banana \\  \\



 Question : \\ \sethlcolor{ao1}\hl{in} \sethlcolor{ao2}\hl{not} \sethlcolor{ao3}\hl{many} \sethlcolor{ao4}\hl{words} \sethlcolor{ao5}\hl{what} \sethlcolor{ao6}\hl{is} \sethlcolor{ao7}\hl{the} \sethlcolor{ao8}\hl{color}
  \sethlcolor{ao9}\hl{of} \sethlcolor{ao10}\hl{this} \sethlcolor{ao11}\hl{fruit}? \\ \\   Predicted Label: \\ 1 \\ \\ \\ \\
    
 Question : \\ \sethlcolor{ct1}\hl{what} \sethlcolor{ct2}\hl{is} \sethlcolor{ct3}\hl{this} ? \\ \\
 Predicted Label: \\ Train \\ \\


 Question : \\ \sethlcolor{at1}\hl{answer} \sethlcolor{at2}\hl{this} \sethlcolor{at3}\hl{for} \sethlcolor{at4}\hl{me} \sethlcolor{at5}\hl{what} \sethlcolor{at6}\hl{is} \sethlcolor{at7}\hl{this} ? \\ \\  Predicted Label:\\ No \\
 
\end{tabular}
\end{center}
\label{tab:im1}
\end{table}

\subsection{Evaluation and Analysis}
The Pythia v3 \cite{singh2019pythia} model achieves an accuracy of 53\% while the Q+I+A model achieves 48.8\% when evaluated on clean samples from the val-set. We tabulate the results obtained by using these phrases as prefixes and suffixes. It is worth noting that when tested on empty questions (which is the baseline for our task) Pythia retains an accuracy of 35.43\% while Q+I+A retains 38.35\%.
Thus our strongest attacks which are meaningful combinations of the basic attacks(in bold; see \autoref{tab:first} for Pythia) and (in bold; see \autoref{tab:third} for Q+I+A) drop the model's accuracy close to the empty question lower bound. Our strongest attack ( see \autoref{tab:first}) renders 97\% of the questions unanswerable, which is a significant increase from 58\% when evaluated on clean questions.

\begin{table}[ht!]

\begin{tabular}{ |l|c|c|}
\hline
\multicolumn{3}{|c|}{Pythia v0.3 \cite{singh2019pythia}} \\
\hline
Prefix Phrase & Accuracy & \% U \\
\hline
 guide me on this & 47.8 & 74.28 \\
 answer this for me &  46.27 & 82.66\\
 in not a lot of words & 44.66 & 85.15 \\
 what is the answer to & 43.46 & 86.10 \\
 in not many words & 42.29 & 91.3 \\
 \hline
 in not many words- & \textbf{38.16} & \textbf{97.06} \\
  what is the answer to & &\\
\hline

\end{tabular}

\caption{Prefix attacks on Pythia v0.3}
\label{tab:first}
\end{table}
\begin{table}[ht!]
\begin{tabular}{ |l|c|c|}
\hline
\multicolumn{3}{|c|}{Pythia v0.3 \cite{singh2019pythia}} \\
\hline
Suffix Phrase & Accuracy & \% U \\
\hline
 guide me on this & 49.8 & 69.2 \\
 answer this for me &  48.82 & 75.19\\
 \hline
 answer this for me- &  45.3 & 82.47\\
 in not a lot of words &  & \\
 answer this for me- &  \textbf{42.5} & \textbf{88.46} \\
 in not many words & & \\
 
\hline
\end{tabular}

\caption{Suffix attacks on Pythia v0.3}
\label{tab:second}
\end{table}
\begin{table}[ht!]
\begin{tabular}{ |l|c|c|}
\hline
\multicolumn{3}{|c|}{Q+I+A \cite{45997}} \\
\hline
Suffix Phrase & Accuracy & \% U  \\
\hline
 describe this for me & 43.52 & 82.8\\
 answer this for me &  43.90 & 89.7\\
 guide me on this & 41.31 & 87.0 \\
 \hline
 answer this for me- & 40.1  & 91.13\\
 in not a lot of words & &\\
 answer this for me- &  \textbf{38.44} & \textbf{94.1}\\
 in not many words & & \\
 
\hline 
\end{tabular}

\caption{Suffix attacks on Q+I+A}
\label{tab:third}
\end{table}
\begin{table}[ht!]
\begin{tabular}{ |l|c|c|}
\hline
\multicolumn{3}{|c|}{Q+I+A \cite{45997}} \\
\hline
Prefix Phrase & Accuracy & \% U  \\
\hline
 describe this for me & 46.72 & 76.8\\
 answer this for me &  45.90 & 79.8\\
 what is the answer to & 44.72 & 80.6\\
 in not many words & 44.50 & 81.4 \\
 \hline
 answer this for me- & \textbf{42.1}  & \textbf{81.13}\\
 in not many words & & \\
\hline 
\end{tabular}

\caption{Prefix attacks on Q+I+A}
\label{tab:fourth}
\end{table}
\section{Performance on other attacks}

\subsection{Word Substitution}
We observe that when we evaluate the model by substituting certain words of the input question by low-attributed words, which change the meaning of the question, the answer predicted in most cases 
is "unanswerable". This means that the model does not over-rely on images and is robust in this aspect.
\subsection{Input Reduction}
We follow the approach of \citealt{feng2018pathologies} to iteratively remove less important words from the input question. With the removal of around 50\% words from a question, the accuracy drops close to 46\% and renders 72\% of the questions unanswerable. The Pythia model is fairly robust in this sense too, as it's output becomes "unanswerable" after considerable input reduction. 

\subsection{Absurd Questions}
To evaluate the effect of absurd attacks on these models, we make a short, non-exhaustive list of objects that do not appear in the validation set of VizWiz(questions, answers and captions) but are present in the training set. We use these objects to form questions similar to the training set questions which contained these objects. A good model should be able to detect absurd questions. For absurd questions like "which country's flag is this ?" (where "flag" does not occur in the validation set of VizWiz) Pythia predicts over 90\% of these (clean image)-(absurd question) pairs as "unanswerable" which is the desired outcome.

\section{Conclusion}
We analyzed two popular VQA models trained under different circumstances for robustness. Our analysis was driven by textual attributions, which helped identify shortcomings of the current approaches to solve a real world problem. The attacks discussed in this paper, illuminate the need for achieving robustness to scale up better to the task of   visual assistance. To improve accessibility for the visually impaired, these VQA systems must be interpretable and safe for operation even under adverse conditions arising out of conversational variations. We believe these insights can be useful to surmount this challenging task.

\bibliography{acl2020}
\bibliographystyle{acl_natbib}

\end{document}